\documentclass[twocolumn]{cinc}
\usepackage{graphicx}
\usepackage{array}

\usepackage[style=ieee,maxbibnames=3]{biblatex}
\begin{document}

\title{Atrial Fibrillation Recurrence Risk Prediction from 12-lead ECG Recorded Pre- and Post-Ablation Procedure}

\author {Eran Zvuloni$^{1}$, Sheina Gendelman$^{1}$, Sanghamitra Mohanty$^{2}$, Jason Lewen$^{3}$, Andrea Natale$^{2}$, \\Joachim A. Behar$^{1}$ \\
\ \\ 
$^1$Faculty of Biomedical Engineering, Technion-IIT, Haifa, Israel\\
$^2$Texas Cardiac Arrhythmia Institute, St David’s Medical Center, Austin, TX, USA\\
$^3$BioSig Technologies Inc.}

\maketitle

\begin{abstract}

\textbf{Introduction:} 12-lead electrocardiogram (ECG) is recorded during atrial fibrillation (AF) catheter ablation procedure (CAP). It is not easy to determine if CAP was successful without a long follow-up assessing for AF recurrence (AFR). Therefore, an AFR risk prediction algorithm could enable a better management of CAP patients. In this research, we extracted features from 12-lead ECG recorded before and after CAP and train an AFR risk prediction machine learning model. \textbf{Methods:} Pre- and post-CAP segments were extracted from 112 patients. The analysis included a signal quality criterion, heart rate variability and morphological biomarkers engineered from the 12-lead ECG (804 features overall). 43 out of the 112 patients ($n$) had AFR clinical endpoint available. These were utilized to assess the feasibility of AFR risk prediction, using either pre or post CAP features. A random forest classifier was trained within a nested cross validation framework. \textbf{Results:} 36 features were found statistically significant for distinguishing between the pre and post surgery states (n=112). For the classification, an area under the receiver operating characteristic (AUROC) curve was reported with $AUROC_{pre}=0.64$ and $AUROC_{post}=0.74$ (n=43). \textbf{Discussion and conclusions:} This preliminary analysis showed the feasibility of AFR risk prediction. Such a model could be used to improve CAP management.

\end{abstract}

\section{Introduction}
Analysis of 12-lead electrocardiogram (ECG) signals is essential for the diagnosis of heart conditions and patient monitoring following intervention. Atrial fibrillation (AF) is the most common arrhythmia with about 3\% prevalence in adults and is associated with 5-fold increase in strokes~\cite{Kirchhof20162016EACTS}. Catheter ablation procedure (CAP) is a treatment for AF ~\cite{Garvanski2019PredictorsAblation}. CAP is considered a long-term success if the patient does not experience AF recurrence (AFR) within a 3-year follow-up~\cite{Garvanski2019PredictorsAblation}. It is estimated that around 70\%~\cite{Kirchhof20162016EACTS} of AF CAPs fail within this timeline. Therefore, there is a need to better understand who the patients are that may best benefit from CAP intervention. Moreover, it is important to closely monitor those who are at high risk of AFR following their CAP. This will support a better management of AF patients.

Engineered features extracted from ECG have been investigated as input for machine learning (ML) algorithms~\cite{Chocron2021RemoteNetwork, 9662857}, supporting complex analysis tasks. These have been providing necessary insights on patients' conditions for risk prediction or diagnosis~\cite{Behar2013ECGReduction, Biton2021AtrialLearning}. A number of research studies used echocardiography for AFR prediction. Yet, there was no agreement on a single echocardiography feature enabling AFR prediction post CAP~\cite{Lizewska-Springer2020EchocardiographicReview}. One work by Fornengo et al~\cite{Fornengo2015PredictionDilation} attempted to harness ML techniques and predicted AFR in cardioversion patients. Their results showed an area under the receiver operating characteristic (AUROC) curve of $0.66$. In an ECG analysis work by Cheng et al~\cite{Cheng2013TheAblation}, the group extracted the f-wave amplitude from three 10-second ECG leads prior to the CAP and analyzed them separately. Two leads were found significant as AFR predictors using the f-wave feature with best result of $Se=0.75$ and $Sp=0.73$.

This research aims to develop a risk prediction model for AFR using features engineered from 12-lead ECG sections taken either before (pre) or after (post) CAP. Accordingly, we attempt to address two fundamental questions: (i) can pre segments classification predict the CAP success rate for a given patient? (ii) can post segments classification answer if a patient treated with CAP is likely to develop AFR? We extracted a large and diverse set of features $(n_x=804)$ from the 12-lead ECG recorded. Then we analyzed the results statistically and within a supervised learning framework. Our results demonstrate the feasibility of predicting AFR on both conditions based on a small dataset of $n=43$ patients.

\section{Methods}
\begin{figure}[h]
\includegraphics[width=7.9cm]{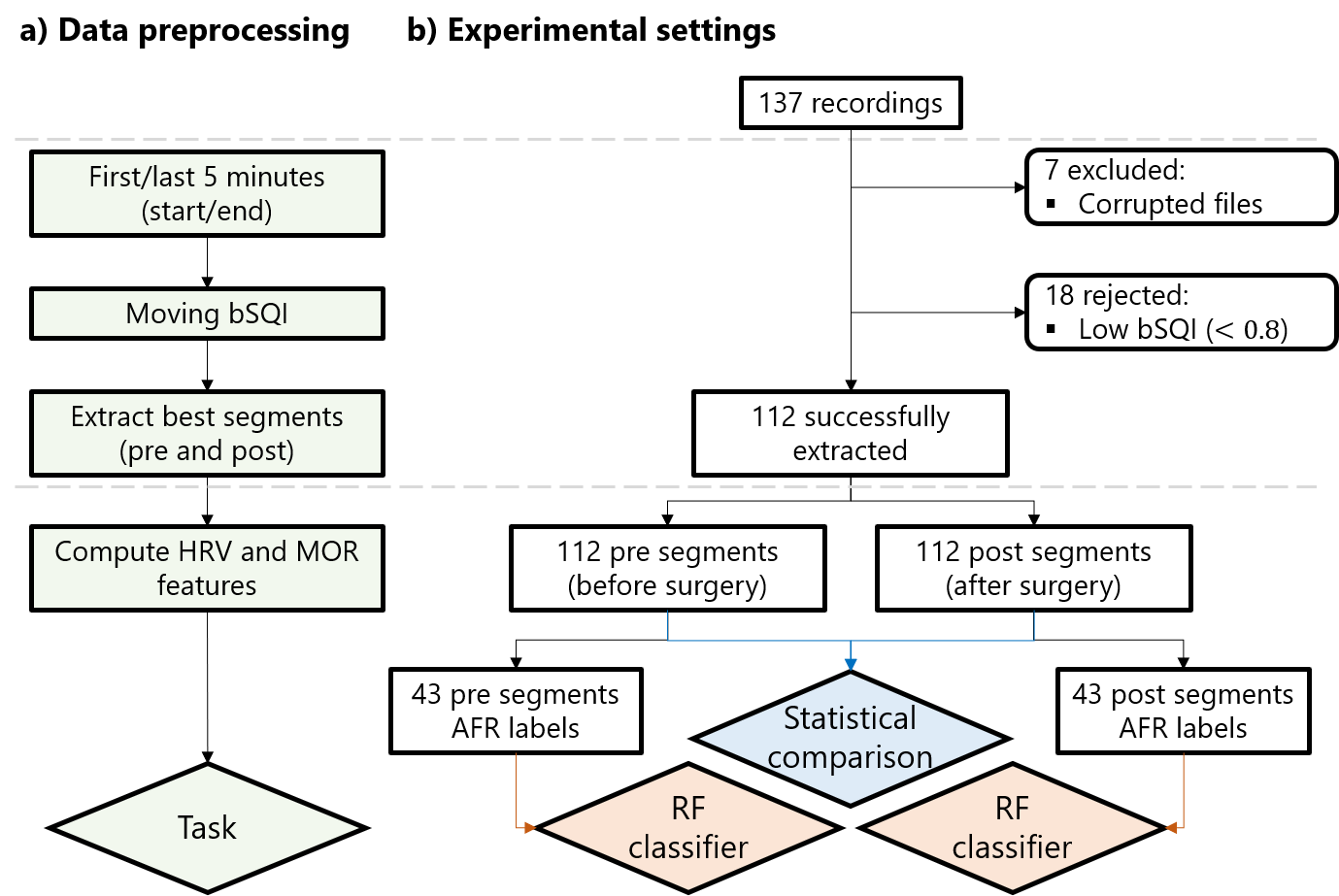}
\caption{Dataset elaboration and experimental setting. a) Data preprocessing and feature engineering: the first and last 5 minutes were scanned to select a pre and post segment(s). Best segments were selected by computing signal quality index (bSQI) in a moving window. Heart rate variability (HRV) and morphological (MOR) features were extracted from the best quality segments. These were used as input for the final tasks (statistical analysis or classification). b) Experimental settings. 43 out of the 112 had atrial fibrillation recurrence (AFR) labels and were classified for AFR prediction using a random forest (RF) model.}
\label{FIG_flow}
\end{figure}

\textbf{\textit{Database and segment extraction:}}
Patients treated with CAP for paroxysmal AF using the PURE EP system (BioSig Technologies Inc.) totaling 137 patients were included in this research. A continuous 12-lead ECG (standard lead system) was recorded throughout the surgery, i.e., starting when the patient received anesthetization before the CAP and until its fading after the treatment. Data was recorded at $f_s=2,000Hz$ with $0.03\mu V$ amplitude quantization and had a median and interquartile duration of $1.6$ and $0.98$ hours, respectively. Fig.~\ref{FIG_flow} describes each recording processing linked to the dataset experimental settings. Seven recordings were corrupted and thus excluded. Then, the first and last 5 minutes from each recording were analyzed. This was intended to reflect the patient state pre and post surgery. Next, representative segments were extracted from each of these 5 minutes. To select the segments with the highest quality, the signals were scanned with a moving signal quality index (bSQI)~\cite{Behar2013ECGReduction}. Since we later applied different statistical and ML tasks, the window had a different size in each task (treated as an hyperparameter), extracting segments with adequate durations: 10 and 60 seconds for the statistical analysis and for the classification tasks, respectively. The bSQI window had 5 seconds of overlapping between the scanned segments. bSQI was computed using two peak detectors, epltd from the WFDB library and jQRS as a reference. Computation was done with the custom PEBM toolbox~\cite{9662857}. Each ECG lead was computed separately and the mean bSQI over the 12 leads was then stored. For each recording, a single segment was extracted for the statistical analysis and the 5 segments with the highest bSQI were extracted for the classification tasks. A total of 18 recordings were discarded because of low quality (bSQI$<0.8$). Among the remaining 112 recordings, 43 had AFR labels available (age$=63.5\pm10.9$; $61.4\%$ males). The labels were based on a follow-up of $320.5\pm120.34$ days with a minimum of $154$ days after the CAP.

\textbf{\textit{Feature engineering:}} 
Two types of ECG features were engineered: heart rate variability (HRV) and morphological (MOR) biomarkers. For the HRV, features 1-20 from Chocron et-al~\cite{Chocron2021RemoteNetwork} with an additional three features denoted “extended parabolic phase space mapping”~\cite{Moharreri2014ExtendedSignal} features were computed. MOR features were computed using the PEBM toolbox~\cite{9662857} for each ECG cycle and the median and standard deviation statistic were computed for each segment. Overall, we obtained 23 HRV and 2×22 MOR features for each lead, thus totaling 804 for the 12 leads. These were used in the statistical analysis. Additional demographics features (META) of age and sex were added to the classification tasks.

\textbf{\textit{Statistical analysis:}}
Statistical significance analysis between the ECG features extracted from the 10-second pre and post highest bSQI segments was performed. A p-value was computed using a paired samples t-test applied on pre-CAP vs post-CAP segment features. Accordingly, each feature obtained a p-value to indicate its significance. Moreover, a mean fold-change (FC) was computed for each feature as
\begin{math}
\mu_{FC}^{feature}=\frac{1}{n}\sum_{i=1}^{n}\frac{f_i^{post}}{f_i^{pre}}
\end{math}
, where $f_i^{pre}$ and $f_i^{post}$ are a feature computed from pre and post segments of patient $i$, respectively, and $n$ is the number of patients.
A volcano plot (Fig.~\ref{FIG_volcano}) was used for display (bioinfokit~\cite{Bedre2021Reneshbedre/bioinfokit:Toolkit}).

\textbf{\textit{Machine learning:}}
A random forest (RF) classifier was trained using the scikit-learn library~\cite{Pedregosa2011Scikit-learn:Python} for the binary classification tasks. Given the low number of recordings having a clinical endpoint available ($n=43$), a nested cross-validation approach was taken. The data was split in a K-fold manner into train, validation and test sets (with no overlap of the recordings), where $K_{test}=8$ for the train-test (outer loop – train includes the validation set) and $K_{val}=8$ for the train-validation (inner loop). A median imputer and a standardizing scaler fitted to the train folds in both loops were used. In addition, the outer loop included minimum redundancy maximum relevance (mRMR) algorithm for feature selection, implemented in MATLAB (Mathworks). In this way the same features were used for all the inner K validations, yet possibly varied between the K test sets. The number of selected features was optimized according to AUROC scores taken from the inner loop. Other hyperparameter optimizations were performed inside the inner loop using a Bayesian search (scikit-optimize) to tune the RF hyperparameters. The search was set to maximize the AUROC. Since we applied data augmentation, the AUROC was computed based on a majority vote over the different segments from a given patient. Moreover, the final AUROC was taken as the mean of the different outer loop 8-fold scores (Fig.~\ref{FIG_ROC}). With this configuration three different models were trained: META, ECG (HRV+MOR), and META+ECG. 

\section{Results}
\begin{figure}[h]
\centering
\includegraphics[width=7.9cm]{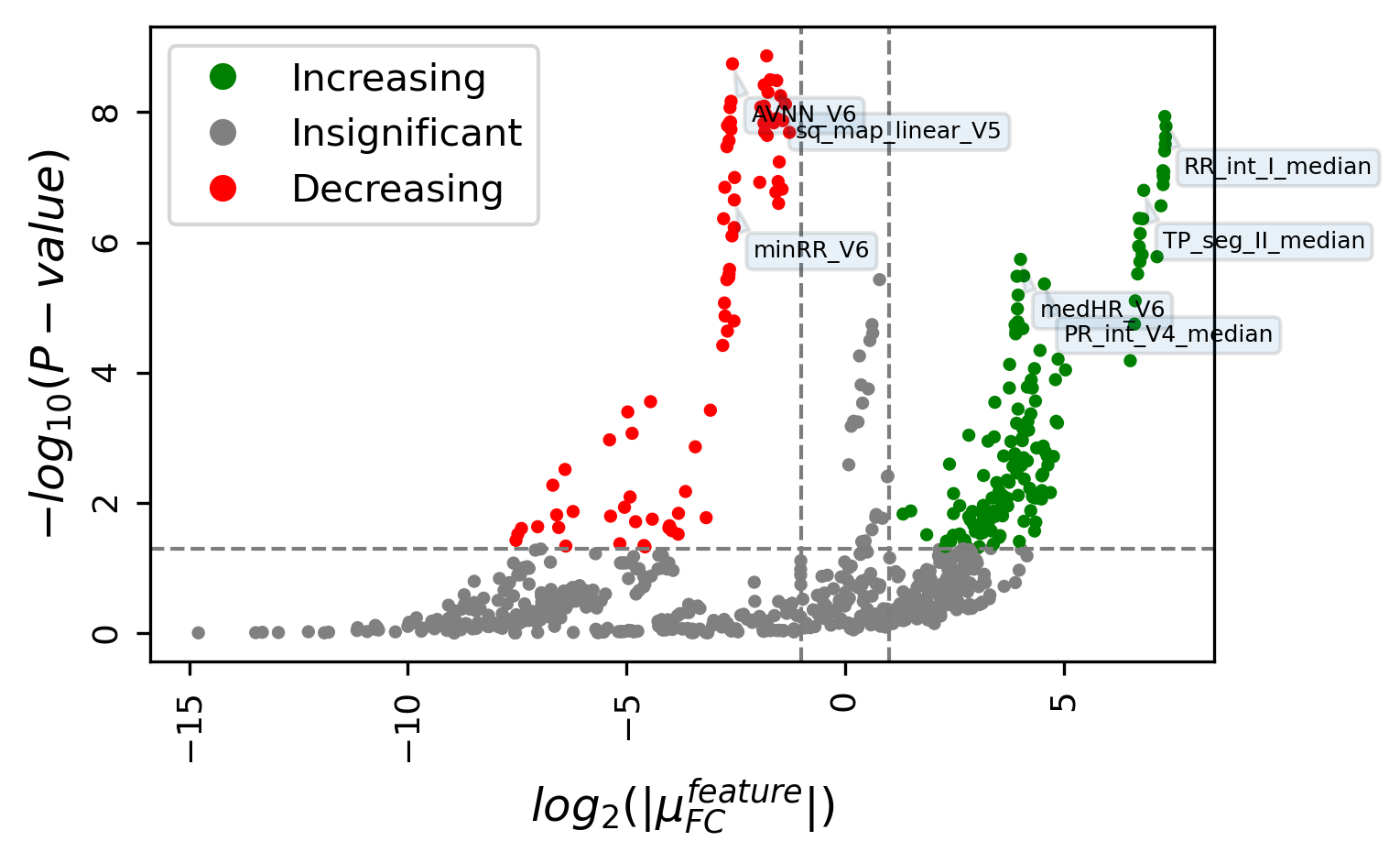}
\caption{Volcano plot showing the engineered features. Features that were found significant (above gray line thresholds) are colored (green and red for an increasing and decreasing fold-change (FC), respectively). The labels show some of these by name and ECG lead.}
\label{FIG_volcano}
\end{figure}

We performed a statistical analysis between the features extracted from the pre and post segments and included the 804 features tested in a volcano plot (Fig.~\ref{FIG_volcano}). The thresholds (gray lines) determine both statistical ($p_{value}^{feature}<\alpha=0.05$) and FC ($|\mu_{FC}^{feature}|>1$) significance. Accordingly, features crossing both thresholds (i.e., closer to the graph top corners) can be considered significant for distinguishing between the patients’ pre- and post-CAP states. With this analysis, we found 36 features to be significant both statistically and FC-wise (Table \ref{tab:sig_feat}).

\begin{table}[]
\small
\caption{\label{tab:sig_feat} Pre- and post-CAP segment statistical analysis.}
\vspace{4 mm}
\begin{tabular}{m{0.8cm} m{6.5cm}}\hline\hline
\textbf{Type} & \textbf{Significant feature found (as in Fig.~\ref{FIG_volcano})}\\
\hline
HRV
&AVNN, IALS, medHR, minRR, PACEv, PAS, PIP, PNN20, PSS, sq\_map\_linear, sq\_map\_intercept\\
\hline
MOR-median 
&$Jpoint$, $PR_{int}$, $PR2_{int}$, $PR_{seg}$, $Pwave_{int}$, $QRS_{int}$, $QT_{int}$, $QT_{cB}$, $R_{dep}$, $RR_{int}$, $R_{wave}$, $TP_{seg}$, $Twave_{int}$, $Twave$\\
\hline
MOR-std 
 &$PR_{int}$, $PR2_{int}$, $Pwave_{int}$, $QRS_{int}$, $QT_{int}$, $QT_{cF}$, $QT_{cH}$, $Rwave$, $ST_{seg}$, $TP_{seg}$, $Twave_{int}$\\
\hline\hline
\end{tabular}
\end{table}

\begin{figure}[h]
\includegraphics[width=8.1cm]{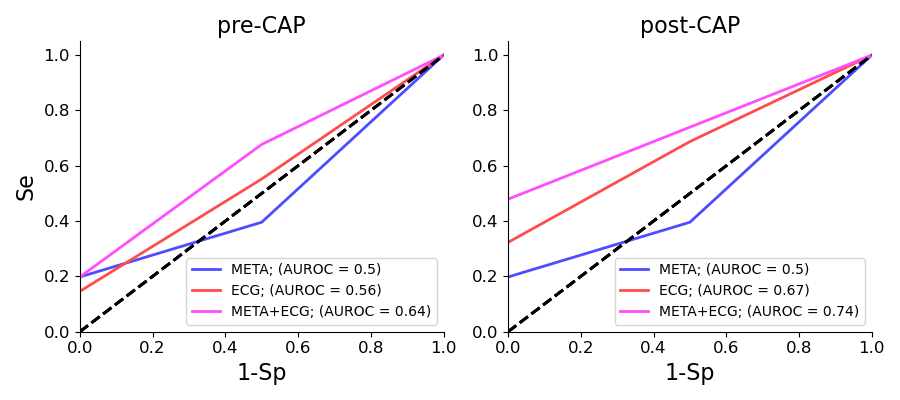}
\caption{Test set receiver operating characteristic curve (ROC) for both RF experiments. Left: experiments obtained by analyzing the ECG acquired from the pre-catheter ablation procedure (CAP) segments. Right: the ECG was analyzed post-CAP.}
\label{FIG_ROC}
\end{figure}

The classification results are shown in Fig.~\ref{FIG_ROC}. mRMR feature selection led to $5$ features being selected. The three models we trained (META, ECG and META+ECG) allowed us to observe the separated and joint effect of the different feature types. Using the META features alone, the RF classifier achieved an $AUROC_{META}=0.5$. In the cases involving the extracted ECG features, the results were  $AUROC_{pre-ECG}=0.6$, $AUROC_{pre-META+ECG}=0.64$, $AUROC_{post-ECG}=0.67$ and $AUROC_{post-META+ECG}=0.74$. 
 
\section{Discussion}
The statistical analysis implies the feasibility of distinguishing between the pre- and post-CAP patients. Importantly, we found significant features constructed from both HRV and MOR analysis (Table \ref{tab:sig_feat}). This feature range demonstrates how conduction is affected by the surgery and may be quantified. Moreover, it emphasizes the importance of combining different feature engineering approaches (e.g., HRV and MOR) and utilizing multiple channels, as all contributed to the separation. Interestingly, the post segments were apparently affected by an increase in heart rate (e.g. using isoproterenol), applied to assess the heart activity post-treatment. This effect may be recognized in the statistical analysis, for example, the median heart rate (medHR) feature was found as a significant discriminator with higher value after surgery (Fig.~\ref{FIG_volcano}).

The ROC curves in Fig.~\ref{FIG_ROC} show a moderate classifications performance for both AFR risk prediction experiments. The extracted ECG features benefit the classification task when compared to using META features alone. Specifically, META had an AUROC of 0.5 versus 0.64 and 0.74 for pre-CAP and post-CAP when using META and ECG features combined. These results match performance reported by others for the task of AFR risk prediction~\cite{Fornengo2015PredictionDilation, Cheng2013TheAblation} using echocardiography or a single ECG lead, although, these research experiments were only used on the pre-treatment 12-lead ECG measurements.

The main limitation of our study is the need to consider the anesthetization effect on patients before and after the CAP. This might have caused bias in the extracted features, which would not correctly reflect the patient state. Thus, ideally, it might be important to acquire ECG data before anesthetization and until enough time has passed post surgery to assume that the drug was washed out. The second main limitation is the low number of patients for which we had an AFR clinical endpoint (only 43), which intrinsically restricts the performance we were able to reach using ML approach.

\section{Conclusions}
HRV and MOR features were extracted from 12-lead ECG recording segments of pre- and post-CAP for AFR treatment. With these we obtained statistically significant separation between patients' pre and post states, implying a heart electrical activity modification caused by the treatment. Moreover, these features were also used to classify between patients that did or did not experience AFR post-treatment (clinical endpoint). The classification showed a moderate AUROC performance for both pre-CAP and post-CAP analysis. Our results serve as a proof of concept and demonstrate how data taken from a 12-lead ECG can be used as a predictor for both treatment success (pre) and arrhythmia likelihood of recurrence (post). In future work, we intend to grow the dataset and to investigate the pre and post differences as features, as well as evaluate deep learning approaches~\cite{Biton2021AtrialLearning} to reach higher performance; thus, allowing the clinical deployment of our model.

\balance

\section*{Acknowledgments}  
EZ, SG and JB acknowledge the support of the Technion EVPR Fund: Hittman Family Fund and BioSig Technologies Inc. This research was partially supported by Israel PBC-VATAT and by the Technion Center for Machine Learning and Intelligent Systems (MLIS).

\printbibliography

@inproceedings{9662857,
    title = {{PhysioZoo ECG: Digital electrocardiography biomarkers to assess cardiac conduction}},
    year = {2021},
    booktitle = {2021 Computing in Cardiology (CinC)},
    author = {Gendelman, Sheina and Biton, Shany and Derman, Raphaël and Zvuloni, Eran and Levy, Jeremy and Lugassy, Snir and Alexandrovich, Alexandra and Behar, Joachim A},
    pages = {1--4},
    volume = {48}
}

@article{Kirchhof20162016EACTS,
    title = {{2016 ESC Guidelines for the management of atrial fibrillation developed in collaboration with EACTS}},
    year = {2016},
    journal = {European Heart Journal},
    author = {Kirchhof, Paulus and Benussi, Stefano and Kotecha, Dipak and Ahlsson, Anders and Atar, Dan and Casadei, Barbara and Castella, Manuel and Diener, Hans-Christoph and Heidbuchel, Hein and Hendriks, Jeroen and Hindricks, Gerhard and Manolis, Antonis S. and Oldgren, Jonas and Popescu, Bogdan Alexandru and Schotten, Ulrich and Van Putte, Bart and Vardas, Panagiotis and Agewall, Stefan and Camm, John and Baron Esquivias, Gonzalo and Budts, Werner and Carerj, Scipione and Casselman, Filip and Coca, Antonio and De Caterina, Raffaele and Deftereos, Spiridon and Dobrev, Dobromir and Ferro, José M. and Filippatos, Gerasimos and Fitzsimons, Donna and Gorenek, Bulent and Guenoun, Maxine and Hohnloser, Stefan H. and Kolh, Philippe and Lip, Gregory Y. H. and Manolis, Athanasios and McMurray, John and Ponikowski, Piotr and Rosenhek, Raphael and Ruschitzka, Frank and Savelieva, Irina and Sharma, Sanjay and Suwalski, Piotr and Tamargo, Juan Luis and Taylor, Clare J. and Van Gelder, Isabelle C. and Voors, Adriaan A. and Windecker, Stephan and Zamorano, Jose Luis and Zeppenfeld, Katja},
    number = {38},
    month = {10},
    pages = {2893--2962},
    volume = {37}
}

@article{Biton2021AtrialLearning,
    title = {{Atrial fibrillation risk prediction from the 12-lead electrocardiogram using digital biomarkers and deep representation learning}},
    year = {2021},
    journal = {European Heart Journal - Digital Health},
    author = {Biton, Shany and Gendelman, Sheina and Ribeiro, Antônio H and Miana, Gabriela and Moreira, Carla and Ribeiro, Antonio Luiz P and Behar, Joachim A},
    month = {8}
}

@article{Behar2013ECGReduction,
    title = {{ECG signal quality during arrhythmia and its application to false alarm reduction}},
    year = {2013},
    journal = {IEEE transactions on biomedical engineering},
    author = {Behar, Joachim and Oster, Julien and Li, Qiao and Clifford, Gari D},
    number = {6},
    pages = {1660--1666},
    volume = {60},
    publisher = {IEEE}
}

@article{Lizewska-Springer2020EchocardiographicReview,
    title = {{Echocardiographic predictors of atrial fibrillation recurrence after catheter ablation: A literature review}},
    year = {2020},
    journal = {Cardiology Journal},
    author = {Lizewska-Springer, Aleksandra and Dabrowska-Kugacka, Alicja and Lewicka, Ewa and Drelich, Lukasz and Krolak, Tomasz and Raczak, Grzegorz},
    number = {6},
    month = {12},
    pages = {848--856},
    volume = {27}
}

@article{Moharreri2014ExtendedSignal,
    title = {{Extended Parabolic Phase Space Mapping (EPPSM): Novel quadratic function for representation of Heart Rate Variability signal}},
    year = {2014},
    journal = {Computing in Cardiology},
    author = {Moharreri, Sadaf and Rezaei, S. and Dabanloo, N. J. and Parvaneh, S.},
    pages = {417--420}
}

@article{Fornengo2015PredictionDilation,
    title = {{Prediction of atrial fibrillation recurrence after cardioversion in patients with left-atrial dilation}},
    year = {2015},
    journal = {European Heart Journal – Cardiovascular Imaging},
    author = {Fornengo, Cristina and Antolini, Marina and Frea, Simone and Gallo, Cristina and Grosso Marra, Walter and Morello, Mara and Gaita, Fiorenzo},
    number = {3},
    month = {3},
    pages = {335--341},
    volume = {16}
}

@article{Garvanski2019PredictorsAblation,
    title = {{Predictors of Recurrence of AF in Patients After Radiofrequency Ablation}},
    year = {2019},
    journal = {European Cardiology Review},
    author = {Garvanski, Iskren and Simova, Iana and Angelkov, Lazar and Matveev, Mikhail},
    number = {3},
    month = {12},
    pages = {165--168},
    volume = {14}
}

@article{Chocron2021RemoteNetwork,
    title = {{Remote Atrial Fibrillation Burden Estimation Using Deep Recurrent Neural Network}},
    year = {2021},
    journal = {IEEE Transactions on Biomedical Engineering},
    author = {Chocron, Armand and Oster, Julien and Biton, Shany and Mandel, Franck and Elbaz, Meyer and Zeevi, Yehoshua Y and Behar, Joachim A},
    number = {8},
    month = {8},
    pages = {2447--2455},
    volume = {68}
}

@misc{Bedre2021Reneshbedre/bioinfokit:Toolkit,
    title = {{reneshbedre/bioinfokit: Bioinformatics data analysis and visualization toolkit}},
    year = {2021},
    author = {Bedre, Renesh},
    month = {1}
}

@article{Pedregosa2011Scikit-learn:Python,
    title = {{Scikit-learn: Machine learning in Python}},
    year = {2011},
    journal = {the Journal of machine Learning research},
    author = {Pedregosa, Fabian and Varoquaux, Gaël and Gramfort, Alexandre and Michel, Vincent and Thirion, Bertrand and Grisel, Olivier and Blondel, Mathieu and Prettenhofer, Peter and Weiss, Ron and Dubourg, Vincent},
    pages = {2825--2830},
    volume = {12},
    publisher = {JMLR. org}
}

@article{Cheng2013TheAblation,
    title = {{The Amplitude of Fibrillatory Waves on Leads aVF and V 1 Predicting the Recurrence of Persistent Atrial Fibrillation Patients Who Underwent Catheter Ablation}},
    year = {2013},
    journal = {Annals of Noninvasive Electrocardiology},
    author = {Cheng, Zhongwei and Deng, Hua and Cheng, Kang'An and Chen, Taibo and Gao, Peng and Yu, Min and Fang, Quan},
    number = {4},
    month = {7},
    pages = {352--358},
    volume = {18}
}

\begin{correspondence}
Dr. Joachim Behar\\
jbehar@technion.ac.il
\end{correspondence}

\end{document}